\documentclass{article}
\usepackage{IEEEtrantools}
\usepackage{multirow}
\usepackage{spconf,amsmath,graphicx}
\usepackage{amsfonts}
\usepackage{verbatim}
\usepackage{setspace}

\title{DIFAI: Diverse Facial Inpainting using StyleGAN Inversion}
\name{Dongsik Yoon$^{\dagger}$\thanks{This research work is supported by DMLab.} \quad Jeong-gi Kwak$^{\dagger}$ \quad Yuanming Li$^{\dagger}$ \quad David Han$^{\ddagger}$ \quad Hanseok Ko$^{\dagger}$}
  
\address{$^{\dagger}$ Korea University  \qquad\qquad $^{\ddagger}$ Drexel Univiersity}
\begin{document}
%
\maketitle
\begin{abstract}
Image inpainting is an old problem in computer vision that restores occluded regions and completes damaged images. In the case of facial image inpainting, most of the methods generate only one result for each masked image, even though there are other reasonable possibilities. To prevent any potential biases and unnatural constraints stemming from generating only one image, we propose a novel framework for diverse facial inpainting exploiting the embedding space of StyleGAN. Our framework employs pSp encoder and SeFa algorithm to identify semantic components of the StyleGAN embeddings and feed them into our proposed SPARN decoder that adopts region normalization for plausible inpainting. We demonstrate that our proposed method outperforms several state-of-the-art methods.
\end{abstract}

\section{Introduction}

 Image inpainting is one of the tasks in computer vision that removes undesired objects or restores occluded regions. Since it is a well-known problem in computer vision, numerous approaches have been proposed in the past. Among them, traditional approaches~\cite{patch3, patch2} propagate small patches from the background area to the missing regions using similarity. However, unlike natural or landscape image inpainting, facial images have unique parts such as nose or mouth, so these methods could not be used. 
 
 With the recent development of generative adversarial learning~\cite{GANs}, GAN-based image inpainting is capable of synthesizing plausible results even when novel objects are present. GLCIC~\cite{glcic} utilized dilated convolutional layers and two auxiliary discriminators in order to complete the missing parts. CA~\cite{CA} was the first attention-based method that uses coarse-to-fine networks for context-aware inpainting. More lately, LBAM~\cite{LBAM} proposed learnable bidirectional attention maps that enable more realistic inpainting for irregular masks. 
 
 For inpainting an image with backgrounds and objects, there are a number of plausible ways to fill in the missing part. Thus, the methods capable of only suggesting one of these many possibilities would have some built-in biases and constraints that may pose limitations on the utilities of such methods. For instance, it would be highly desirable in GAN-based data augmentation to have a method capable of suggesting a variety of possibilities in a missing region of an image.

 To overcome this limitation, PIC~\cite{PIC} used a short+long term attention layer to synthesize pluralistic inpainting results. Furthermore, this method proposed a probabilistic principled framework comprised of two parallel paths called generative path and reconstructive path. PD-GAN~\cite{PDGAN} provides diverse inpainting results using the proposed SPDNorm Resblock. PD-GAN also adds perceptual diversity loss for various inpainting possibilities in the training process. Contrary to the well-known diversity loss~\cite{divloss} perceptual diversity loss is calculated on the perceptual space for keeping the context unchanged and training stable. 
 
 On a related front, numerous researches in GAN have been conducted to synthesize diverse high-resolution images. StyleGAN-based methods~\cite{stylegan1, stylegan2, stylegan3} have been highly successful in generating some astonishing images by controlling the latent space. Pixel2style2pixel (pSp)~\cite{psp} proposed an encoder network that directly generates a sequence of style vectors which are input into a StyleGAN decoder, forming the extended latent space. SeFa~\cite{sefa} or GANSpace~\cite{GANSPACE} uncover relevant directions in the latent space of pre-trained StyleGAN that affect the semantic properties of the decoded image in an unsupervised manner. The aforementioned methods are capable of generating a variety of images in a range of details from coarse features such as shapes or poses to fine properties such as lighting, background attributes, or feature variations. 

\begin{figure*}[!t]
\centering
\includegraphics[scale=0.53]{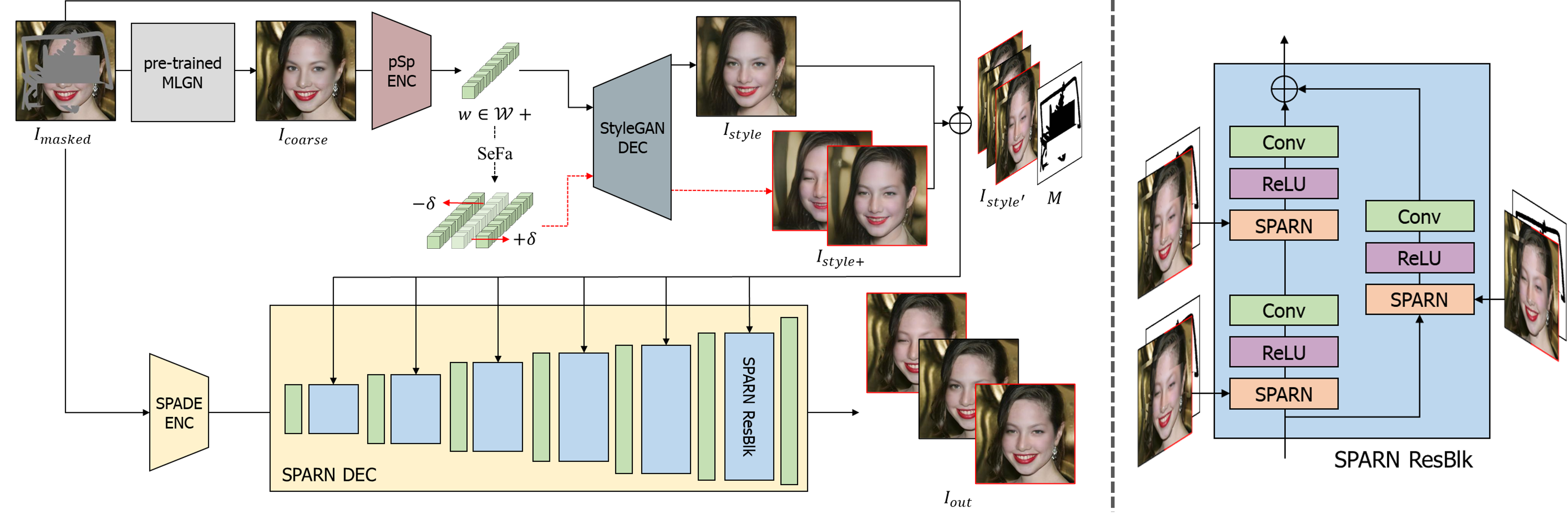}
\vspace{-3mm}
\caption{Summary of the proposed framework. $I_{style^{\prime}}$ and $M$ are the input of SPARN decoder. In the SPARN decoder, each region normalization layer uses the $I_{style^{\prime}}$ and $M$ to modulate the layer activations.}
\vspace{-3mm}
\label{fig1}
\end{figure*}

 Given this, we propose a novel framework for diverse facial inpainting based on controlling StyleGAN's latent space for generating a set of plausible inpainted regions while maintaining the remaining regions. Our approach requires only an image with a masked region as input. Our framework first coarsely completes the input (masked) image from the pre-trained inpainting network so that pSp encoder will extract style vectors in the latent space. Afterward, we manipulate the latent space in meaningful directions to transform the semantic attributes of the decoded images. By feeding the manipulated latent space into the StyleGAN decoder, we could generate images with transformed facial shapes or attributes. Additionally, we feed decoded images as a condition into our proposed spatially adaptive region normalization (SPARN) decoder. 
 
 Our proposed decoder adopts region normalization~\cite{RN} in each layer to allow synthesizing realistic inpainting results. Thus, the proposed generator can be trained to perform more diverse image inpainting using StyleGAN without any prior condition. Based on our experiments of the publicly released dataset CelebA-HQ, we demonstrate that the proposed approach delivers superior performance compared to various state-of-the-art approaches specialized in inpainting tasks.

\section{Proposed Method}
\subsection{Proposed Framework}
This section introduces our proposed facial image inpainting framework. As shown in Fig.~\ref{fig1}, our framework consists of four parts: a pre-trained inpainting network, a pSp encoder, a StyleGAN decoder, and the proposed generator. We first apply a customized MLGN~\cite{MLGN} model for coarse inpainting. Our customized MLGN differs from the original model by an adjustment made to the lambda parameter for synthesizing blurry results. The blurry results would promote more diverse embeddings by pSp encoder, thereby allowing StyleGAN to generate more diverse image inpainting. We generate ground truth and masked image pairs as 

\begin{equation}
I_{masked}=I_{gt} \odot M,
\end{equation}
where $I_{gt}$ is the ground truth image, $M$ is the mask applied to erase portions of the ground truth image, and $I_{masked}$ is the masked image. We input $I_{masked}$ into a pre-trained customized MLGN as

\begin{equation}
I_{coarse}=\text{cMLGN}(I_{masked}).
\end{equation}

\begin{figure*}[!t]
\centering
\begin{tabular}{@{\,}c@{\,}@{\,}c@{\,}c@{\,}c@{\,}c@{\,}c@{\,}c@{\,}c@{\,}c@{\,}}
\includegraphics[width=1.9cm]{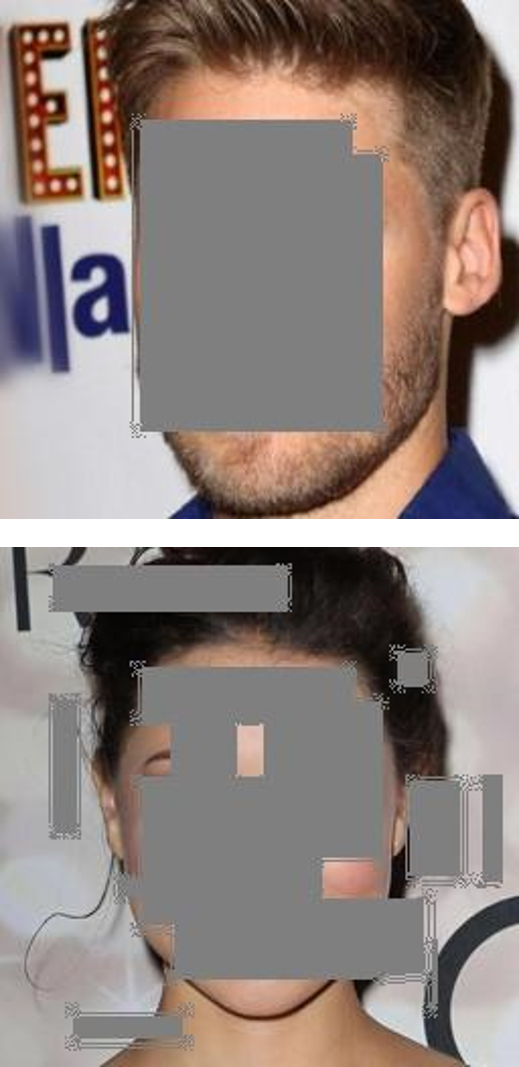}&
\includegraphics[width=1.9cm]{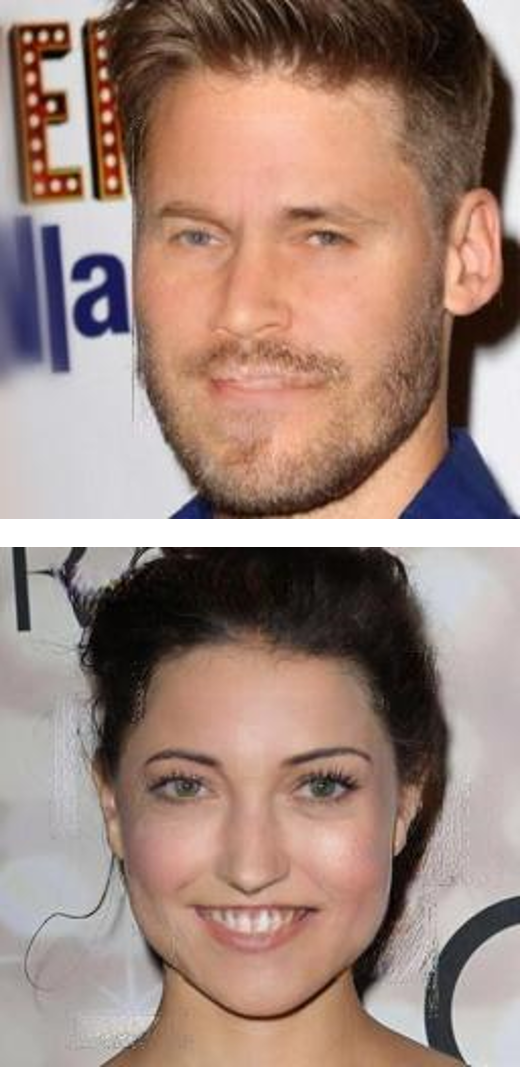}&
\includegraphics[width=1.9cm]{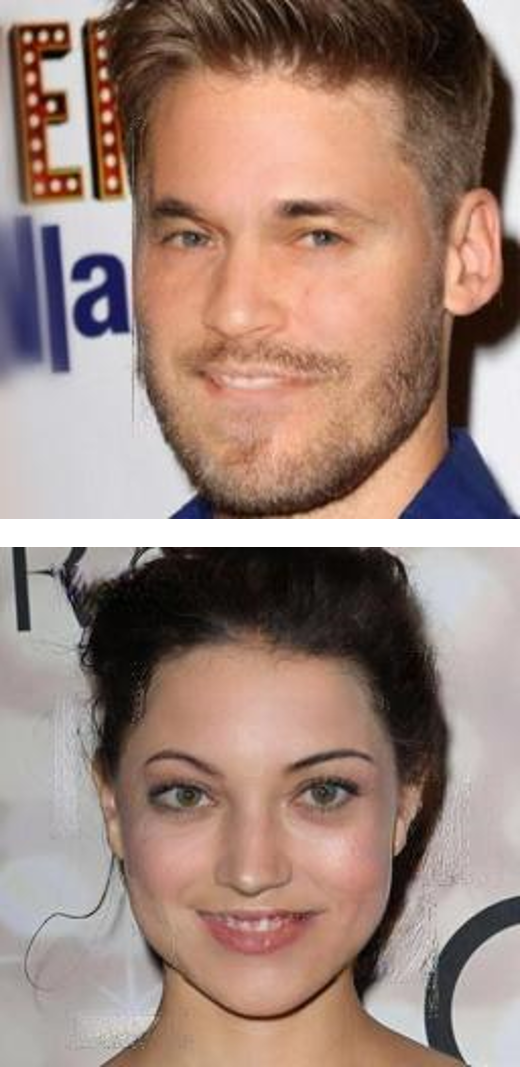}&
\includegraphics[width=1.9cm]{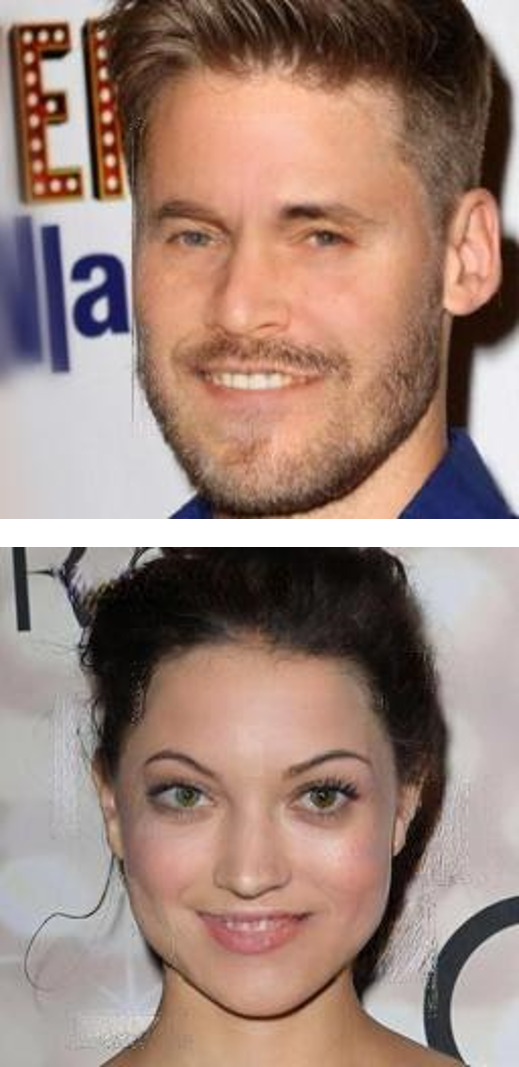}&
\includegraphics[width=1.9cm]{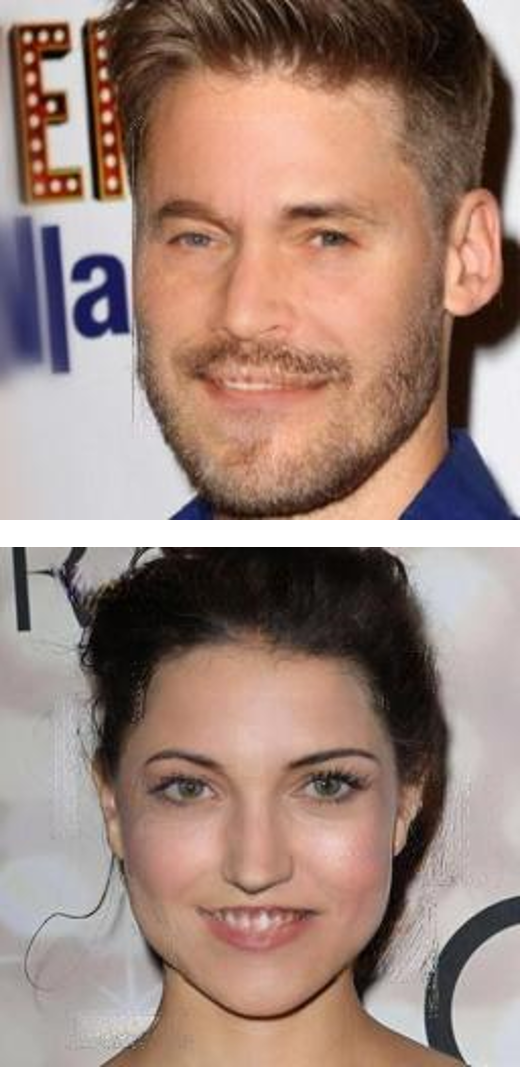}&
\includegraphics[width=1.9cm]{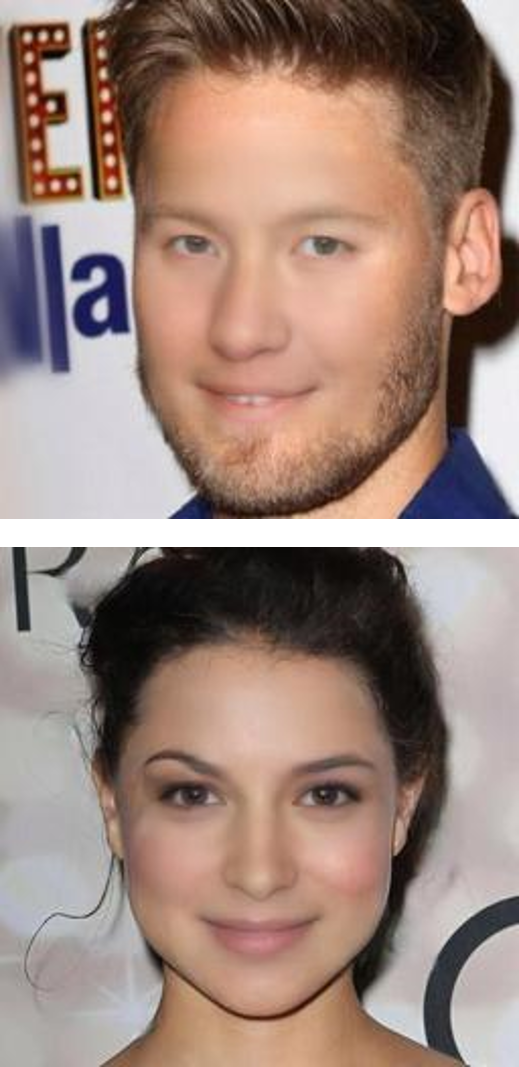}&
\includegraphics[width=1.9cm]{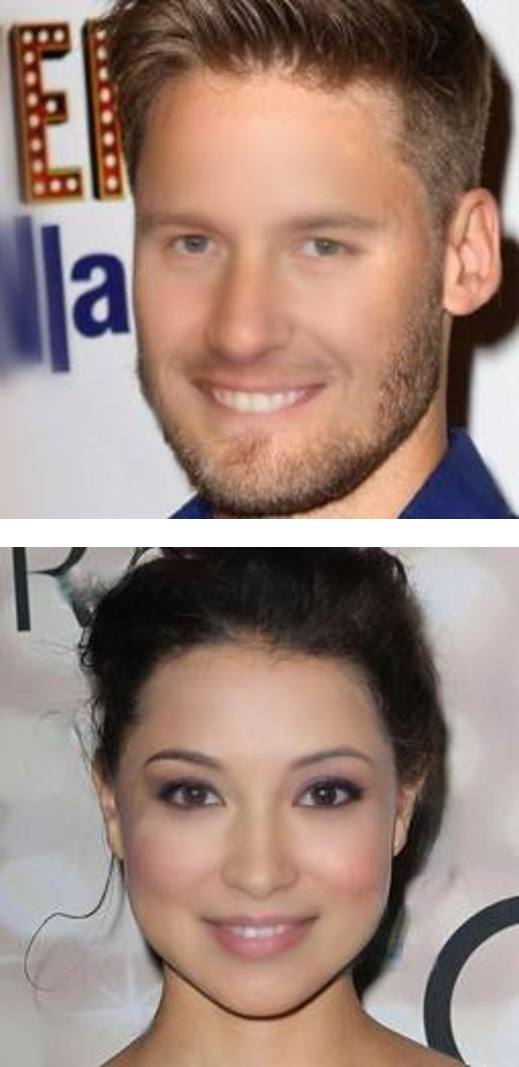}&
\includegraphics[width=1.9cm]{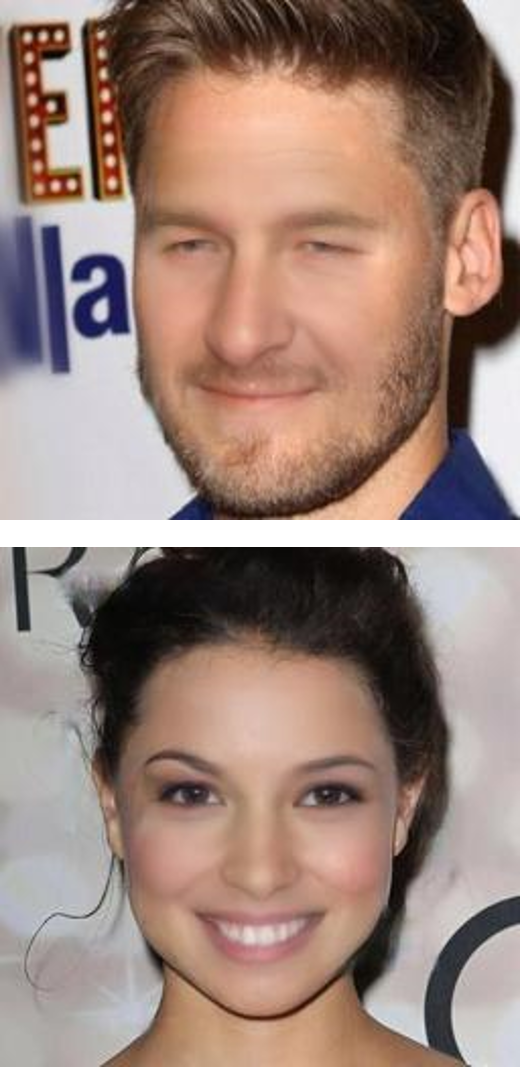}&
\includegraphics[width=1.9cm]{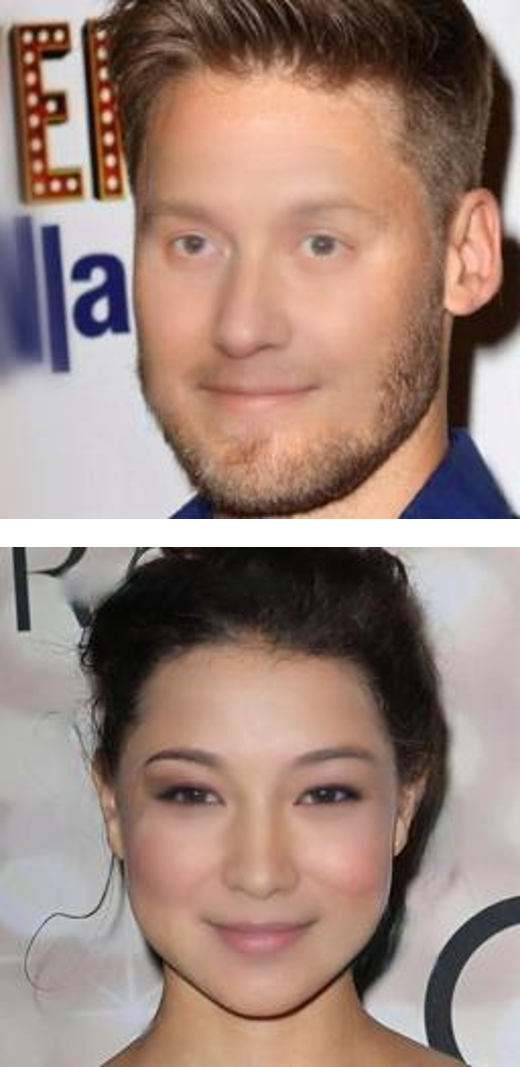}
\\
Input&$\text{PIC}_{1}$\cite{PIC}&$\text{PIC}_{2}$&$\text{PIC}_{3}$&$\text{PIC}_{4}$&$\text{Ours}_{1}$&$\text{Ours}_{2}$&$\text{Ours}_{3}$&$\text{Ours}_{4}$
\end{tabular}
\\
\vspace{-1mm}
\caption{Pluralistic qualitative comparison with PIC~\cite{PIC} and Ours.}
\vspace{-3mm}
\label{fig2}
\end{figure*}

\begin{figure}[]
\centering
\begin{tabular}{@{\,}c@{\,}@{\,}c@{\,}c@{\,}c@{\,}c@{\,}c@{\,}c@{\,}c@{\,}c@{\,}}
\includegraphics[width=1.65cm]{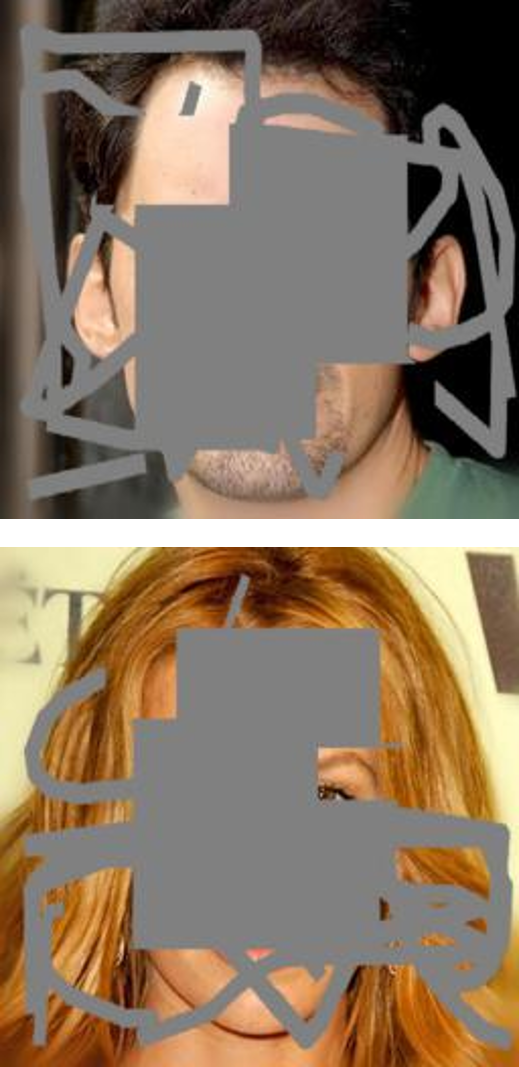}&
\includegraphics[width=1.65cm]{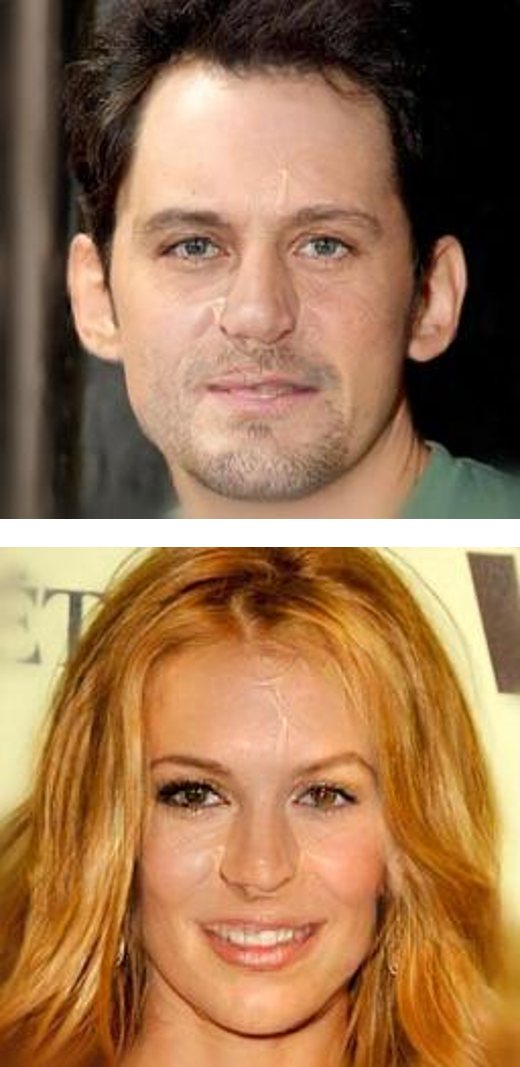}&
\includegraphics[width=1.65cm]{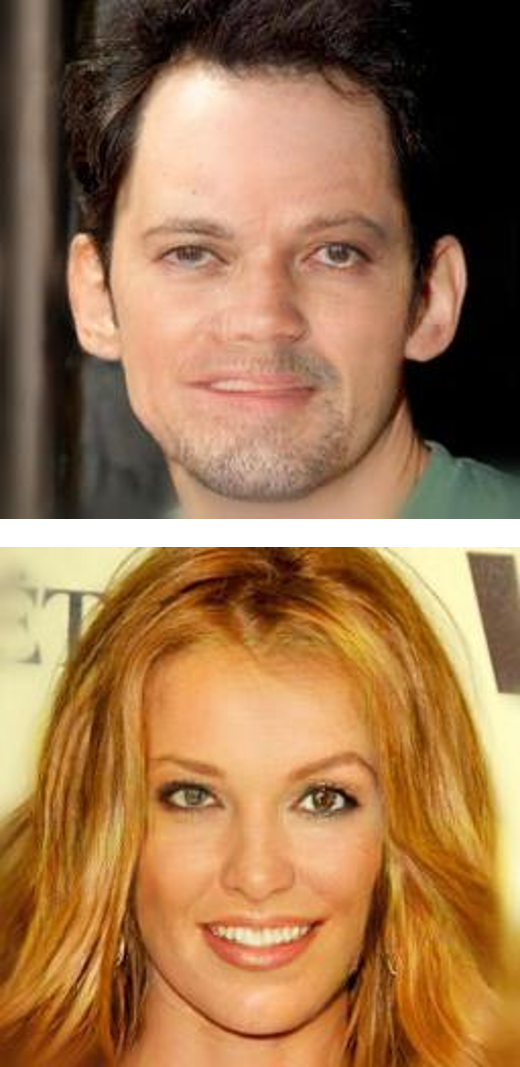}&
\includegraphics[width=1.65cm]{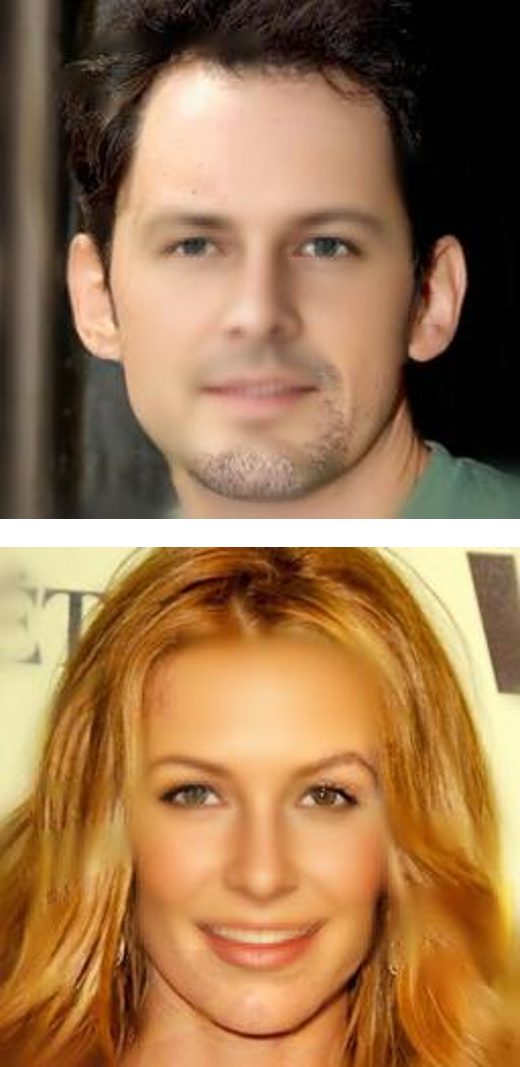}&
\includegraphics[width=1.65cm]{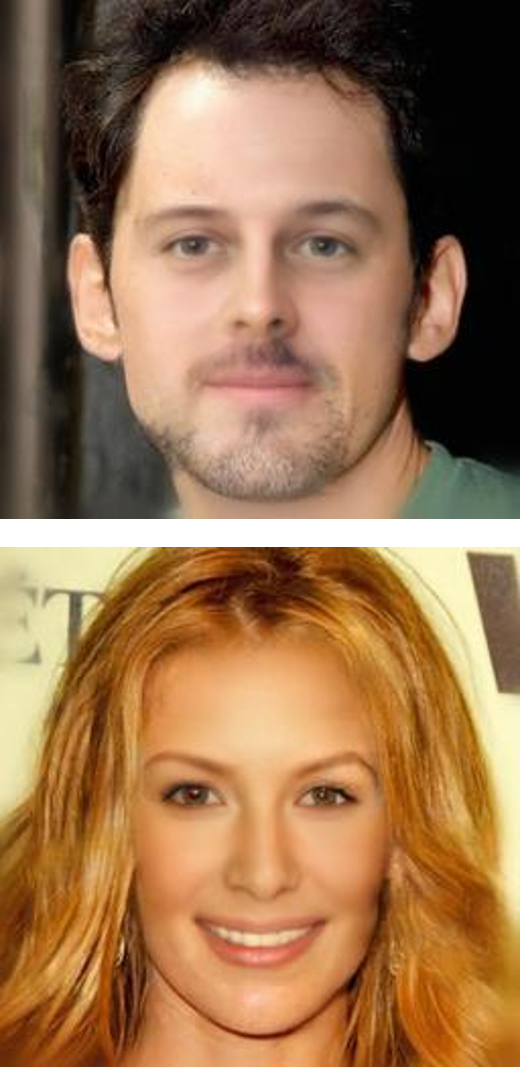}&
\\
Input&LBAM\cite{LBAM}&EC\cite{EC}&MLGN\cite{MLGN}&Ours
\end{tabular}
\\
\vspace{-1mm}
\caption{Qualitative comparison with LBAM~\cite{LBAM}, EC~\cite{EC}, MLGN~\cite{MLGN}, and Ours.}
\label{fig3}
\end{figure}

\begin{figure*}
\centering
\includegraphics[scale=0.6]{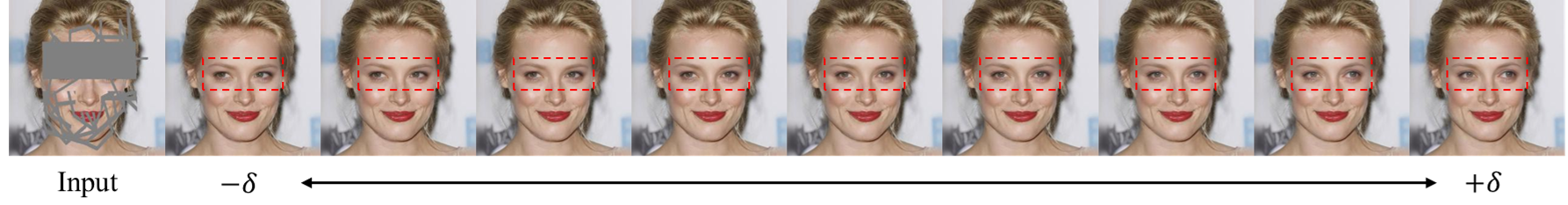}
\vspace{-8mm}
\caption{Illustration of continuous inpainting results using SeFa that adjusts delta gradually.}
\vspace{-3mm}
\label{fig4}
\end{figure*}

Most image inpainting methods generate only one result for each masked image, even though there are many other possibilities. As such, there are always possibilities of unrealistic biases and constraints due to the network being forced to produce only one of many plausible results. 
To prevent such artificial biases, we use StyleGAN-based~\cite{stylegan2} image augmentation that is capable of synthesizing a variety of images that has a similar structure as the ground truth but with changed facial attributes. $I_{coarse}$ is applied to pSp~\cite{psp} encoder which maps the embedding vector $w$ in a latent space $W+$. The extracted $w$ is then decoded to produce an initial set of diverse images using StyleGAN. 
Afterward, we use the SeFa~\cite{sefa} algorithm that performs eigen-decomposition of StyleGAN's weight matrix to discover principal components that span dominant changes in the decoded images.
We feed the proposed generator with the embeddings $w_{\delta_i}$ perturbed by $\delta_{i}$ in a number of principal directions during the training process for synthesizing multiple images.

\begin{align}
\begin{split}
I_{style}&=\text{StyleGAN}(\text{pSp}(I_{corase})), \\
I_{style+}&=\{ \text{StyleGAN}(w_{\delta_{1}}), ..., \text{StyleGAN}(w_{\delta_{\alpha}})\},
\end{split}
\end{align}
where $\alpha$ is the number of $I_{style+}$ for training.
Our generator $G(\cdot)$ is comprised of a SPADE~\cite{SPADE} encoder and the proposed SPARN decoder. The proposed SPARN decoder may look similar to the existing SPADE decoder, but it can maintain consistency in the masked and unmasked regions by using region normalization~\cite{RN} for image inpainting. 
As an input condition of SPARN decoder, we define $I_{style^{\prime}}$ as,

\begin{equation}
I_{style^{\prime}}=I_{masked}+M_{r}\odot \{ I_{style}, I_{style+}\}, 
\end{equation}
where, $M_{r}$ is the reversed mask.
Additionally, we input $I_{masked}$ into the SPADE encoder to ensure that features present in the masked image are maintained in the output images. Our SPARN decoder consists of SPARN residual block and following upsampling layers. Since each residual block runs at a different scale, we downsample the input $M$ and $I_{style^{\prime}}$ to match the spatial resolution. Thereby, we could perform more diverse facial image inpainting using as conditions various images that transformed several facial attribute detail.

\begin{equation}
I_{out}=G(I_{masked}, I_{style^{\prime}}, M)
\end{equation}

Furthermore, we encode $I_{out}$ and $I_{gt}$ into the discriminator for calculating adversarial loss. We use Spectral Normalization~\cite{spec1, spec2} in the discriminator since it is faster and more stable than other normalizations by a simple formulation.

\subsection{Loss Function}
In order to synthesize plausible and realistic image inpainting, we define our loss function in two parts: inpainting loss and adversarial loss. Our proposed inpainting loss is composed of four components: reconstruction loss, VGG style loss, perceptual loss, and MS-SSIM loss. 
Reconstruction loss completes occluded regions using $l1$-norm error. By comparing the generated image to the ground truth, we calculate the hole region loss and valid region loss, respectively. Additionally, we define perceptual loss and VGG style loss with the VGG-19 network~\cite{vgg19} pre-trained on ImageNet. As the name indicates, perceptual loss~\cite{perceploss} measures the feature map distance between the generated image and the ground truth image. Our perceptual loss looks similar to the aforementioned perceptual loss, but we further measure the distance between $I_{out}$ with $I_{style}$. We adopt $M_{r}$ to reflect $I_{style}$ more plausible in erased regions. We defined our perceptual loss as,

\begin{IEEEeqnarray}{CC}
L_{per}=\sum_{i}\sum_{j=1}^{\alpha}||F_{i}(I_{out_{j}})\cdot M_{r}-F_{i}(I_{style_{j}})\cdot M_{r}||_{1} \nonumber\\
\qquad\qquad+\sum_{i}||F_{i}(I_{out})\cdot M-F_{i}(I_{gt})\cdot M||_{1},
\end{IEEEeqnarray}
where $F_{i}$ denotes the feature maps of the $i\,'th$ layer of a VGG-19 network. We use VGG style loss, as defined by ~\cite{styleloss}, which alleviates “checkerboard” artifacts caused by upsampling convolution layers. Our VGG style loss also compares $I_{out}$ with $I_{style}$, using $M_{r}$.

\begin{IEEEeqnarray}{CC}
L_{style}=\sum_{k}\sum_{j=1}^{\alpha}||G^{F}_{k}(I_{out_{j}})\cdot M_{r}-G^{F}_{k}(I_{style_{j}})\cdot M_{r}||_{1} \nonumber\\
\qquad\qquad+\sum_{k}|G^{F}_{k}(I_{out})\cdot M-G^{F}_{k}(I_{gt})\cdot M||_{1},
\end{IEEEeqnarray}
where $G^{F}_{k}$ is a gram matrix consisting of feature maps $F_{k}$. 
Additionally, we customize another loss function by utilizing MS-SSIM~\cite{MSSSIM, MLGN}, which is one of the image quality comparison approaches.

\begin{equation}
L_{\text{MS-SSIM}}=1-\frac{1}{N}\sum_{i=1}^{N}\text{MS-SSIM}_{n}
\end{equation}

We calculate adversarial loss using WGAN-GP which optimizes the Wasserstein distance. 
We define adversarial loss $L_{G}$ and $L_{D}$ as,
\begin{IEEEeqnarray}{CCL}
 L_{G}=\mathbb{E}_{I_{masked}}[D(G(I_{masked}, I_{style^{\prime}}, M))], \\
 L_{D}=\mathbb{E}_{I_{gt}}[D(I_{gt})]-\mathbb{E}_{I_{out}}[D(I_{out})]\nonumber\\
 \qquad\qquad\qquad-\lambda_{gp}\mathbb{E}_{\hat{I}}[(||	\nabla_{\hat{I}}D(\hat{I})||_{2}-1)^{2}].
\end{IEEEeqnarray}

Our overall loss denoted as,
\begin{IEEEeqnarray}{CC}
L_{all}=\lambda_{adv}L_{adv}+\lambda_{ssim}L_{\text{MS-SSIM}}+\lambda_{sty}L_{style}\nonumber\\
\qquad\qquad+L_{per}+\lambda_{hole}L_{hole}+\lambda_{valid}L_{valid},
\end{IEEEeqnarray}
where, $\lambda$ are hyper-parameters that control the terms' relative importance.

\begin{table}[]
\begin{center}
\begin{tabular}{l|c|c|c|c|c}
                                           & Mask      & Ours          & LBAM  & EC    & MLGN  \\ \hline\hline
\multicolumn{1}{c|}{\multirow{5}{*}{SSIM}} & Quickdraw & \textbf{0.833} & 0.821 & 0.817 & 0.832 \\ \cline{2-6} 
\multicolumn{1}{c|}{}                      & 10-20\%   & 0.837          & 0.814 & 0.827 &  \textbf{0.839} \\ \cline{2-6} 
\multicolumn{1}{c|}{}                      & 20-30\%   &  \textbf{0.777}          & 0.744 & 0.761 &  \textbf{0.777} \\ \cline{2-6} 
\multicolumn{1}{c|}{}                      & 30-40\%   &  \textbf{0.709}          & 0.667 & 0.681 & 0.706 \\ \cline{2-6} 
\multicolumn{1}{c|}{}                      & 40-50\%   &  \textbf{0.633}          & 0.583 & 0.595 & 0.624 \\ \hline\hline
\multirow{5}{*}{LPIPS}                     & Quickdraw & 0.049           &  \textbf{0.047} &  \textbf{0.047} & 0.063 \\ \cline{2-6} 
                                           & 10-20\%   & 0.052           & 0.057 &  \textbf{0.051} & 0.065 \\ \cline{2-6} 
                                           & 20-30\%   & 0.084           & 0.087 &  \textbf{0.082} & 0.103 \\ \cline{2-6} 
                                           & 30-40\%   &  \textbf{0.124}           & 0.126 & 0.128 & 0.148 \\ \cline{2-6} 
                                           & 40-50\%   &  \textbf{0.170}           & 0.172 & 0.183 & 0.201 \\ \hline\hline
\multirow{5}{*}{FID}                       & Quickdraw & 25.95            &  \textbf{25.79} & 27.49 & 28.45 \\ \cline{2-6} 
                                           & 10-20\%   &  \textbf{24.74}            & 27.63 & 25.65 & 26.73 \\ \cline{2-6} 
                                           & 20-30\%   &  \textbf{34.55}            & 36.51 & 34.80 & 38.34 \\ \cline{2-6} 
                                           & 30-40\%   &  \textbf{46.87}            & 48.47 & 47.14 & 52.54 \\ \cline{2-6} 
                                           & 40-50\%   & 64.86            &  64.40 & \textbf{63.75} & 73.07 \\ \hline
\end{tabular}
\end{center}
\vspace{-1mm}
\caption{Quantitative comparison on CelebA-HQ. In each row, the best results are shown in bold text.}
\vspace{-3mm}
\label{table1}
\end{table}

\section{Experiments}
\subsection{Implement Details}

For the implementation, we used the Pytorch library. Our hyper-parameters $\lambda_{adv}$, $\lambda_{ssim}$, $\lambda_{sty}$, $\lambda_{hole}$, and $\lambda_{valid}$ are set to 0.5, 120, 3, and 0.5 respectively. 
In this paper, we evaluate all the models using CelebA-HQ dataset and split them into two groups: 28,000 selected for training and 2000 for testing. We used 256×256 images with irregular holes to train and evaluate the proposed methods. In addition, we combine Quickdraw irregular mask dataset~\cite{irrmask} with 85×85 square holes in random positions to create more irregular holes. By combining square holes with the Quickdraw dataset, the model becomes more robust to irregular holes.

\begin{table}[]
\begin{center}
\begin{tabular}{c|c|c|c}
\multicolumn{1}{l|}{}                                                        & Mask    & PIC    & Ours            \\ \hline\hline
\multirow{2}{*}{\begin{tabular}[c]{@{}c@{}}Diversity\\ (LPIPS) $\uparrow$ \end{tabular}} & 20-30\% & 0.0714 & \textbf{0.0849} \\ \cline{2-4} 
                                                                             & 30-40\% & 0.1079 & \textbf{0.1313} \\ \hline
\end{tabular}
\vspace{-1mm}
\caption{Quantitative comparison of diversity on CelebA-HQ. For the diversity comparison, higher LPIPS is better.}
\vspace{-5mm}
\label{table2}
\end{center}
\end{table}

\subsection{Qualitative Comparisons}

First, we compare the image inpainting quality of our baseline against three state-of-the-art methods. 
Fig.~\ref{fig3} describes the images generated by the proposed method and those generated by the other methods. 
Our model is superior to all the others in the aspect of image quality and plausibility. Fig.~\ref{fig2} compares diverse images generated by PIC~\cite{PIC} and ours. Compares to the PIC, our method accomplishes more diverse and pluralistic instances.

\begin{table}[]
\begin{center}
\begin{tabular}{c|c|c|c}
      & \begin{tabular}[c]{@{}c@{}}SPADE\end{tabular} & w/o RN & Ours           \\ \hline\hline
SSIM  & 0.816                                                   & 0.817  & \textbf{0.833} \\ \hline
LPIPS & 0.056                                                   & 0.057  & \textbf{0.049} \\ \hline
FID   & 27.96                                                   & 28.45  & \textbf{25.95} \\ \hline
\end{tabular}
\vspace{-1mm}
\caption{Quantitative comparison of ablation study on CelebA-HQ. We choose the Quickdraw irregular mask here.}
\vspace{-5mm}
\label{table3}
\end{center}
\end{table}

\subsection{Quantitative Comparisons}
We implement a quantitative comparison of image inpainting to three existing methods and our own, using different types and sizes of masks. As shown in Table.~\ref{table1}, our method outperforms three metrics SSIM, LPIPS~\cite{LPIPS}, and FID~\cite{FID} to existing methods that specialize in only image inpainting tasks.
Table.~\ref{table2} shows that our method achieves a relatively higher diversity score than another method. The diversity score is calculated between 4K pairs synthesized from a sampling of 1K images.
Overall, we feed the $w$ into StyleGAN to calculate Table.~\ref{table1} and feed the $w_{\delta}$ to calculate the diversity score.

To justify the effectiveness of the proposed SPARN decoder, we conduct the ablation study as follows: 1) Using SPADE~\cite{SPADE} decoder; 2) replacing all the region normalization~\cite{RN} with batch normalization (w/o RN). 
As shown in Table.~\ref{table3}, each of the proposed sub-modules performs a very important role in the overall architecture.


\section{Conclusion}
We propose a novel method of generating diverse facial inpainted images based on manipulating StyleGAN embedding space. To properly discover meaningful direction and the associated variations for diverse facial inpaintings, we utilize pSp encoder and SeFa algorithm. These embedded vectors and the variations are then fed into our proposed SPARN decoder as conditions for diverse inpainting. 
From the proposed framework, we demonstrated that our method synthesizes plausible diverse images from a single masked input while maintaining high inpainting quality. 

\vfill\pagebreak

\bibliographystyle{IEEEbib}
\bibliography{strings,refs}

\begin{thebibliography}{10}

\bibitem{patch3}
Alexei~A Efros and Thomas~K Leung,
\newblock ``Texture synthesis by non-parametric sampling,''
\newblock in {\em ICCV}, 1999.

\bibitem{patch2}
Connelly Barnes, Eli Shechtman, Adam Finkelstein, and Dan~B Goldman,
\newblock ``Patchmatch: A randomized correspondence algorithm for structural
  image editing,''
\newblock {\em ToG}, 2009.

\bibitem{GANs}
Ian Goodfellow, Jean Pouget-Abadie, Mehdi Mirza, Bing Xu, David Warde-Farley,
  Sherjil Ozair, Aaron Courville, and Yoshua Bengio,
\newblock ``Generative adversarial nets,''
\newblock in {\em NeurIPS}, 2014.

\bibitem{glcic}
Satoshi Iizuka, Edgar Simo-Serra, and Hiroshi Ishikawa,
\newblock ``Globally and locally consistent image completion,''
\newblock {\em ToG}, 2017.

\bibitem{CA}
Jiahui Yu, Zhe Lin, Jimei Yang, Xiaohui Shen, Xin Lu, and Thomas~S Huang,
\newblock ``Generative image inpainting with contextual attention,''
\newblock in {\em CVPR}, 2018.

\bibitem{LBAM}
Chaohao Xie, Shaohui Liu, Chao Li, Ming-Ming Cheng, Wangmeng Zuo, Xiao Liu,
  Shilei Wen, and Errui Ding,
\newblock ``Image inpainting with learnable bidirectional attention maps,''
\newblock in {\em ICCV}, 2019.

\bibitem{PIC}
Chuanxia Zheng, Tat-Jen Cham, and Jianfei Cai,
\newblock ``Pluralistic image completion,''
\newblock in {\em CVPR}, 2019.

\bibitem{PDGAN}
Hongyu Liu, Ziyu Wan, Wei Huang, Yibing Song, Xintong Han, and Jing Liao,
\newblock ``Pd-gan: Probabilistic diverse gan for image inpainting,''
\newblock in {\em CVPR}, 2021.

\bibitem{divloss}
Qi~Mao, Hsin-Ying Lee, Hung-Yu Tseng, Siwei Ma, and Ming-Hsuan Yang,
\newblock ``Mode seeking generative adversarial networks for diverse image
  synthesis,''
\newblock in {\em CVPR}, 2019.

\bibitem{stylegan1}
Tero Karras, Samuli Laine, and Timo Aila,
\newblock ``A style-based generator architecture for generative adversarial
  networks,''
\newblock in {\em CVPR}, 2019.

\bibitem{stylegan2}
Tero Karras, Samuli Laine, Miika Aittala, Janne Hellsten, Jaakko Lehtinen, and
  Timo Aila,
\newblock ``Analyzing and improving the image quality of stylegan,''
\newblock in {\em CVPR}, 2020.

\bibitem{stylegan3}
Tero Karras, Miika Aittala, Samuli Laine, Erik H\"ark\"onen, Janne Hellsten,
  Jaakko Lehtinen, and Timo Aila,
\newblock ``Alias-free generative adversarial networks,''
\newblock in {\em NeurIPS}, 2021.

\bibitem{psp}
Elad Richardson, Yuval Alaluf, Or~Patashnik, Yotam Nitzan, Yaniv Azar, Stav
  Shapiro, and Daniel Cohen-Or,
\newblock ``Encoding in style: a stylegan encoder for image-to-image
  translation,''
\newblock in {\em CVPR}, 2021.

\bibitem{sefa}
Yujun Shen and Bolei Zhou,
\newblock ``Closed-form factorization of latent semantics in gans,''
\newblock in {\em CVPR}, 2021.

\bibitem{GANSPACE}
Erik H{\"a}rk{\"o}nen, Aaron Hertzmann, Jaakko Lehtinen, and Sylvain Paris,
\newblock ``Ganspace: Discovering interpretable gan controls,''
\newblock {\em Arxiv}, 2020.

\bibitem{RN}
Tao Yu, Zongyu Guo, Xin Jin, Shilin Wu, Zhibo Chen, Weiping Li, Zhizheng Zhang,
  and Sen Liu,
\newblock ``Region normalization for image inpainting,''
\newblock in {\em AAAI}, 2020.

\bibitem{MLGN}
Jie Liu and Cheolkon Jung,
\newblock ``Facial image inpainting using multi-level generative network,''
\newblock in {\em ICME}, 2019.

\bibitem{EC}
Kamyar Nazeri, Eric Ng, Tony Joseph, Faisal Qureshi, and Mehran Ebrahimi,
\newblock ``Edgeconnect: Structure guided image inpainting using edge
  prediction,''
\newblock in {\em ICCVW}, 2019.

\bibitem{SPADE}
Taesung Park, Ming-Yu Liu, Ting-Chun Wang, and Jun-Yan Zhu,
\newblock ``Semantic image synthesis with spatially-adaptive normalization,''
\newblock in {\em CVPR}, 2019.

\bibitem{spec1}
Takeru Miyato, Toshiki Kataoka, Masanori Koyama, and Yuichi Yoshida,
\newblock ``Spectral normalization for generative adversarial networks,''
\newblock {\em Arxiv}, 2018.

\bibitem{spec2}
Jiahui Yu, Zhe Lin, Jimei Yang, Xiaohui Shen, Xin Lu, and Thomas~S Huang,
\newblock ``Free-form image inpainting with gated convolution,''
\newblock in {\em ICCV}, 2019.

\bibitem{vgg19}
Karen Simonyan and Andrew Zisserman,
\newblock ``Very deep convolutional networks for large-scale image
  recognition,''
\newblock {\em Arxiv}, 2014.

\bibitem{perceploss}
Justin Johnson, Alexandre Alahi, and Li~Fei-Fei,
\newblock ``Perceptual losses for real-time style transfer and
  super-resolution,''
\newblock in {\em ECCV}, 2016.

\bibitem{styleloss}
Mehdi~SM Sajjadi, Bernhard Scholkopf, and Michael Hirsch,
\newblock ``Enhancenet: Single image super-resolution through automated texture
  synthesis,''
\newblock in {\em ICCV}, 2017.

\bibitem{MSSSIM}
Zhou Wang, Eero~P Simoncelli, and Alan~C Bovik,
\newblock ``Multiscale structural similarity for image quality assessment,''
\newblock in {\em ACSSC}, 2003.

\bibitem{irrmask}
Karim Iskakov,
\newblock ``Semi-parametric image inpainting,''
\newblock {\em Arxiv}, 2018.

\bibitem{LPIPS}
Richard Zhang, Phillip Isola, Alexei~A Efros, Eli Shechtman, and Oliver Wang,
\newblock ``The unreasonable effectiveness of deep features as a perceptual
  metric,''
\newblock in {\em CVPR}, 2018.

\bibitem{FID}
Martin Heusel, Hubert Ramsauer, Thomas Unterthiner, Bernhard Nessler, and Sepp
  Hochreiter,
\newblock ``Gans trained by a two time-scale update rule converge to a local
  nash equilibrium,''
\newblock in {\em NeurIPS}, 2017.

\end{thebibliography}

\end{document}